\documentclass[sigconf]{acmart}
\AtBeginDocument{%
  \providecommand\BibTeX{{%
    \normalfont B\kern-0.5em{\scshape i\kern-0.25em b}\kern-0.8em\TeX}}}

\usepackage{amssymb}
\usepackage{amsmath,amsfonts}
\usepackage{bm}
\usepackage{algorithmic}
\usepackage{algorithm}
\usepackage{array}
\usepackage[caption=false,font=normalsize,labelfont=sf,textfont=sf]{subfig}
\usepackage{textcomp}
\usepackage{stfloats}
\usepackage{url}
\usepackage{verbatim}
\usepackage{graphicx}
\usepackage{multirow}
\usepackage{threeparttable}
\usepackage{booktabs}
\usepackage{bookmark}
\usepackage{graphicx}
\usepackage{dsfont}
\begin{document}

\title{PEVA-Net: Prompt-Enhanced View Aggregation Network for Zero/Few-Shot Multi-View 3D Shape Recognition}

\author{Dongyun Lin, Yi Cheng, Shangbo Mao, Aiyuan Guo, Yiqun Li}
% \authornote{Dongyun Lin}
% \author{Yi Cheng}

% \author{Shangbo Mao}
\affiliation{%
  \institution{Institute for Infocomm Research, A*STAR}
  \streetaddress{1 Fusionopolis Way, $\#$21-01 Connexis, Singapore 138632}
  \country{Singapore}
  \postcode{138632}
}

% \author{Anonymous Authors}
%% You do not have to enter your paper ID


\begin{abstract}
  Large vision-language models have impressively promote the performance of 2D visual recognition under zero/few-shot scenarios. In this paper, we focus on exploiting the large vision-language model, i.e., CLIP, to address zero/few-shot 3D shape recognition based on multi-view representations. The key challenge for both tasks is to generate a discriminative descriptor of the 3D shape represented by multiple view images under the scenarios of either without explicit training (zero-shot 3D shape recognition) or training with a limited number of data (few-shot 3D shape recognition). We analyze that both tasks are relevant and can be considered simultaneously. Specifically, leveraging the descriptor which is effective for zero-shot inference to guide the tuning of the aggregated descriptor under the few-shot training can significantly improve the few-shot learning efficacy. Hence, we propose Prompt-Enhanced View Aggregation Network (PEVA-Net) to simultaneously address zero/few-shot 3D shape recognition. Under the zero-shot scenario, we propose to leverage the prompts built up from candidate categories to enhance the aggregation process of multiple view-associated visual features. The resulting aggregated feature serves for effective zero-shot recognition of the 3D shapes. Under the few-shot scenario, we first exploit a transformer encoder to aggregate the view-associated visual features into a global descriptor. To tune the encoder, together with the main classification loss, we propose a self-distillation scheme via a feature distillation loss by treating the zero-shot descriptor as the guidance signal for the few-shot descriptor. This scheme can significantly enhance the few-shot learning efficacy.

  Without any pre-training process, our PEVA-Net can produce the state-of-the-art zero-shot 3D shape recognition performance on ModelNet40, ModelNet10 and ShapeNetCore 55 datasets with the accuracy of \textbf{84.48\%}, \textbf{93.50\%} and \textbf{74.65\%}. Under the 16-shot setting of ModelNet40, the proposed PEVA-Net also sets the state-of-the-art recognition accuracy of \textbf{90.64\%}. Extensive ablation experiments are conducted to analyze the superiority of the proposed PEVA-Net.
\end{abstract}

\ccsdesc[500]{Computing methodologies~Shape inference}
\ccsdesc[500]{Computing methodologies~Shape representations}

\keywords{Multi-View 3D Shape Recognition, Vision-Language Model, CLIP, Zero-Shot Learning, Few-Shot Learning}



\maketitle

\section{Introduction}
Along with the substantial development of 3D sensory technology, a huge amount of 3D data are generated to support many industry applications such as virtual reality, autonomous driving and mechanical part design and inspection~\cite{qi2021review}. Towards an effective management of large-scale 3D data, one fundamental task is to accurately identify the categories of 3D shapes.

In the community of 3D computer vision and multimedia, substantial research efforts have been made to address 3D shape recognition based on various 3D representations, such as point clouds, voxels and multi-view images. Thanks to the impressive development of deep learning in 2D computer vision~\cite{he2016deep,ren2016faster}, multi-view based methods for 3D shape recognition produce the state-of-the-art performance~\cite{lin2023multi,wei2022learning,su2015multi,lin2022multi}. However, the training of deep learning based methods requires a huge number of data which are extremely laborious to collect and conduct annotation. Hence, it is in demand to develop a 3D shape recognition system which can achieve high accuracy with the lowest or even no training resources. To this end, many research works, e.g., ~\cite{zhang2022pointclip,zhu2023pointclip,huang2023clip2point}, are conducted on zero-shot and few-shot 3D shape recognition where none or an extremely limited number of training samples are provided.

Recently, large vision-language models have impressively promote the performance of zero/few-shot 2D vision tasks. Particularly, Contrastive Language-Image Pretraining (CLIP)~\cite{radford2021learning} and its variants~\cite{mu2022slip,cherti2023reproducible} leverage nature language as the supervision to guide the visual feature learning in a contrastive learning manner. The learned visual features are highly generalized and transferable to address the zero-shot visual tasks, such as image classification~\cite{radford2021learning}, anomaly detection~\cite{jeong2023winclip} and video understanding~\cite{Xu2021VideoCLIPCP}. In 3D vision, CLIP is also widely exploited for zero-shot and few-shot 3D shape recognition~\cite{huang2023clip2point,Hegde2023CLIPG3,xue2023ulip,xue2023ulip,liu2024openshape,qi2023contrast,zhang2022pointclip,zhu2023pointclip,shen2024diffclip,song2023mv}. A group of works, e.g., PointCLIPs~\cite{zhang2022pointclip,zhu2023pointclip} and CLIP2Point~\cite{huang2023clip2point}, projected 3D point clouds into the mutli-view depth maps and then leveraged the CLIP's text and visual encoders to conduct zero-shot inference. The bottleneck of these methods is the large domain gap between the projected depth images of point clouds and the images of nature scenes which are heavily collected to train CLIP. To alleviate this limitation, DiffCLIP~\cite{shen2024diffclip} was proposed to exploit a diffusion model to transfer the style of depth maps to that of the nature scene images. The recently published MV-CLIP~\cite{song2023mv} was proposed to exploit CLIP for zero-shot 3D shape recognition based on the multi-view rendered images of 3D meshes since the rendered images occupy the smaller domain gap with the nature scene images. MV-CLIP modified both visual and text paths of CLIP via adopting view selection and hierarchical prompts to achieve the state-of-the-art recognition accuracy on zero-shot 3D shape recognition.

\begin{figure}[!t]
  \centering
  % \vspace{8mm}
  \includegraphics[scale=0.25]{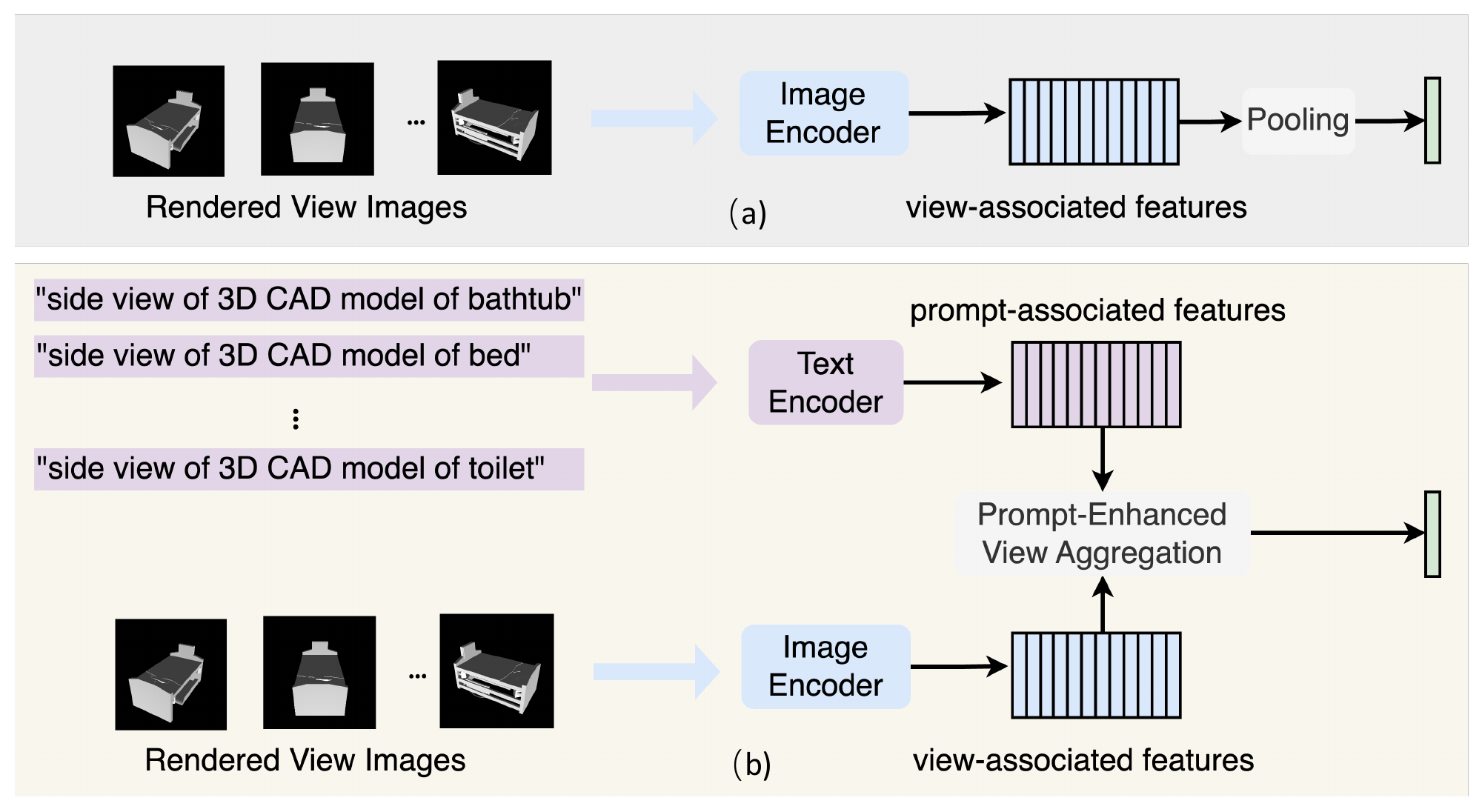}
  \caption{Comparison between (a) the trivial view aggregation scheme via pooling; (b) our proposed prompt-enhanced view aggregation by leveraging the prompt-associated and the view-associated features.}
  \label{fig:comparison_ideas}
\end{figure}

In this paper, we also focus on exploiting CLIP to address zero/few-shot 3D shape recognition based on the multi-view rendered images of 3D shapes. The core challenge for both tasks is how to generate a discriminative descriptor of the 3D shape by effectively aggregating the view images under the scenarios of either without any explicit training (zero-shot) or training with a limited number of data (few-shot). For zero-shot 3D shape recognition, the trivial solution to aggregate the multi-view visual features is average or max pooling across the view-associated visual features as illustrated in Fig.~\ref{fig:comparison_ideas}(a). However, the pooling operation overlooks the difference in the discriminative capability of views since they are captured from varied perspectives. To overcome this limitation, we propose a prompt-enhance view aggregation scheme (illustrated in Fig.~\ref{fig:comparison_ideas}(b)) to aggregate view-associated features based on the prompt-associated features. Specifically, we leverage the prompts built up from the candidate categories to determine a view-specific discriminative score which can quantify the discriminative capability of each view image. The view-associated visual features are aggregated via a weighted summation based these discriminative scores. The resulting aggregated feature serves for effective zero-shot recognition of the 3D shapes. Since our network adopts the prompt information to enhance the view aggregation process, it is named as Prompt-Enhanced View Aggregation Network (PEVA-Net).

% \begin{figure}[htbp]
%   \centering
%   % \vspace{8mm}
%   \includegraphics[scale=0.29]{weight_view_visualization.pdf}
%   \caption{Examples of view images of 3D shapes with their discriminative scores. For simplicity, for each 3D shape, we only show the view images with the lowest and the highest discriminative scores, respectively. The examples are picked from the testing set of ModelNet40.}
%   \label{fig:weight_view_visualization}
% \end{figure}

\begin{figure}[htbp]
  \centering
  % \vspace{8mm}
  \includegraphics[scale=0.35]{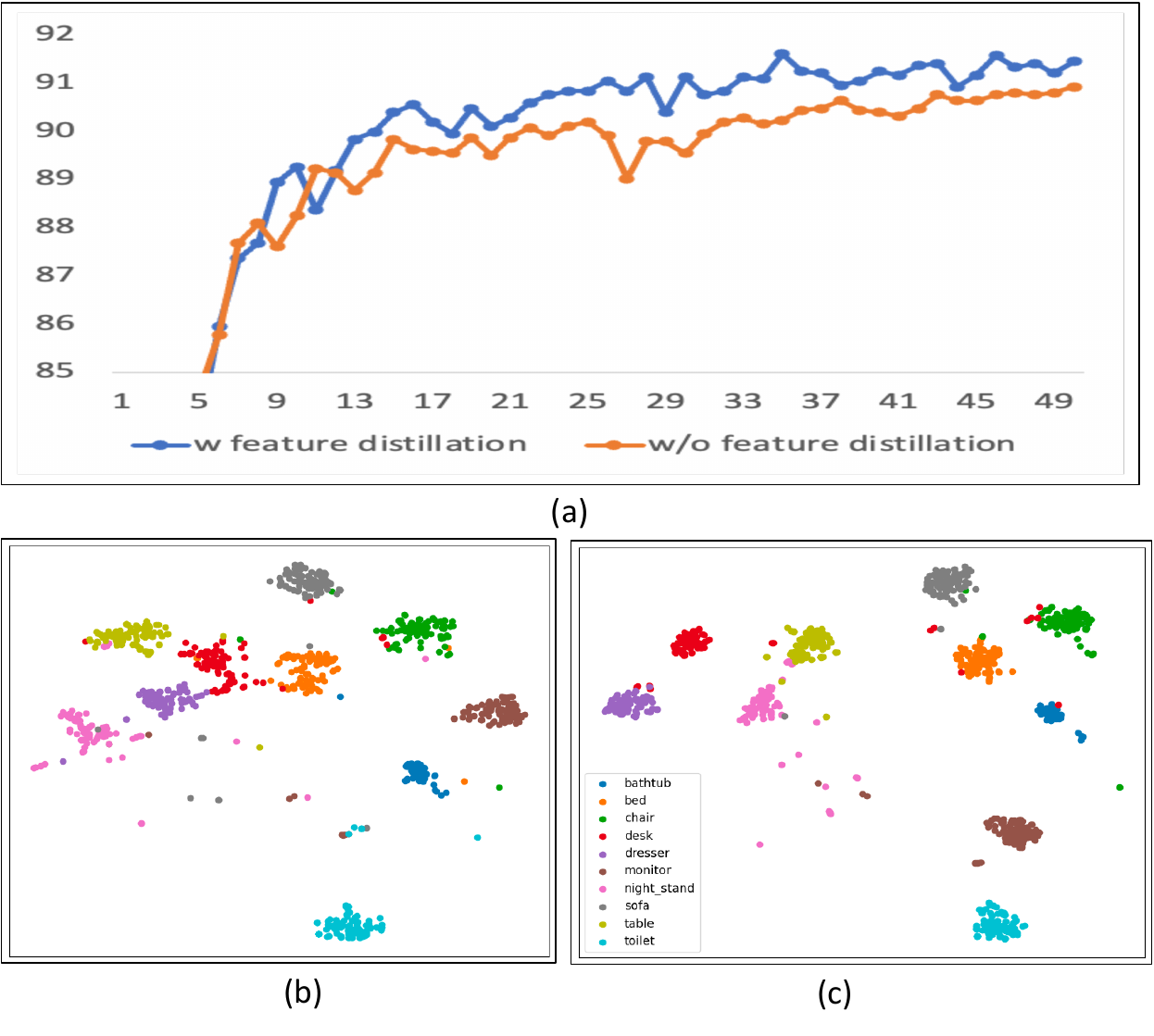}
  \caption{The empirical observations to show the effectiveness of the proposed self-distillation scheme: (a) The recognition accuracy on ModelNet40 test set across the training epochs under 16-shot setting; (b) and (c): 2D t-SNE embeddings produced by PEVA-Net without feature distillation and PEVA-Net with feature distillation, respectively, on the testing samples from 10 categories.}
  \label{fig:model_acc_visualization_and_tsne}
\end{figure}

Under the few-shot scenario, our PEVA-Net first exploits a transformer encoder to aggregate the view-associated visual features into a global descriptor. To alleviate overfitting due to only a few training samples are provided, we propose a self-distillation scheme by leveraging the descriptor which is effective for zero-shot inference to guide the training of the aggregated descriptor under the few-shot condition via feature distillation. Specifically, to tune the encoder, we propose a training loss consisting of two components: (i) the main classification loss based on the logits calculated as the inner product between the prompt-associated features and the few-shot descriptor; and (ii) the auxiliary feature distillation loss by treating the zero-shot descriptor as the guidance signal for the few-shot descriptor. The effectiveness of feature distillation from the zero-shot descriptor to the few-shot descriptor can be empirically investigated by scrutinizing the training process and the learned feature embeddings as shown in Fig.~\ref{fig:model_acc_visualization_and_tsne}. From Fig.~\ref{fig:model_acc_visualization_and_tsne}(a), it is observed that PEVA-Net trained with feature distillation can clearly outperform that without feature distillation in term of recognition accuracy across the training epochs. From Fig.~\ref{fig:model_acc_visualization_and_tsne}(b) and Fig.~\ref{fig:model_acc_visualization_and_tsne}(c), it is noted that PEVA-Net trained with feature distillation is capable of generating more separable feature embeddings than its counterpart trained without feature distillation. These empirically observations show the effectiveness of the proposed self-distillation scheme.

% (shown in Fig.~\ref{fig:model_acc_visualization_and_tsne}(a)), where under the 16-shot setting (16 3D shapes are picked up from each category for training), the model trained with feature distillation can clearly outperform the model without feature distillation. Furthermore, we also plot the 2D t-SNE embeddings produced by our PEVA-Net without feature distillation (shown in Fig.~\ref{fig:model_acc_visualization_and_tsne}(b)) and by that with feature distillation (shown in Fig.~\ref{fig:model_acc_visualization_and_tsne}(c)). It is noted that the model training with feature distillation is capable of generating more separable and discriminative embeddings. These observations show the effectiveness of the proposed self-distillation scheme via the feature distillation loss.

% The empirical observations to show the effectiveness of the proposed self-distillation scheme:

To summarize, we propose PEVA-Net based on CLIP to address zero-shot and few-shot multi-view 3D shape recognition. The major contributions of this paper are summarized below:

\begin{itemize}

   \item We propose a prompt-enhanced view aggregation module to leverage the prompts built up from the candidate categories to enhance the aggregation process of the view-associated visual features for effective zero-shot 3D shape recognition.

   \item We propose a self-distillation scheme by leveraging the zero-shot descriptor to guide the training of the few-shot descriptor via feature distillation to significantly improve the few-shot learning efficacy.

   % descriptor effective for zero-shot inference to guide the training of the descriptor under few-shot setting via feature distillation to significantly improve the few-shot learning efficacy.

   \item Extensive experiments conducted on ModelNet40, ModelNet10 and ShapeNetCore 55 demonstrate the proposed PEVA-Net could achieve the state-of-the-art performance on zero-shot and few-shot multi-view based 3D shape recognition.

   % The proposed PEVA-Net could achieve the state-of-the-art performance on zero-shot and few-shot multi-view based 3D shape recognition on the datasets of ModelNet40, ModelNet10 and ShapeNetCore 55.

\end{itemize}

\section{Related Work}
In this section, the related works on multi-view based 3D shape recognition and zero/few-shot 3D shape recognition are reviewed.
\subsection{Multi-view Based 3D Shape Recognition}
Multi-view based methods address 3D shape recognition represented by multiple 2D view images captured by the cameras from different perspectives. Motivated by the success of 2D CNN in deep feature learning~\cite{he2016deep,krizhevsky2012imagenet}, multi-view based methods employing CNN as the view-associated feature extractor have produced the state-of-the-art performance in the supervised 3D shape recognition~\cite{li2019angular,Chen2021MVTMV,lin2023multi,wei2022learning,su2015multi,lin2022multi}. The pioneering work multi-view convolutional neural network (MVCNN)~\cite{su2015multi} proposed to aggregate the CNN extracted view features via maximum pooling for 3D shape recognition. Under this paradigm, many works were proposed to improve MCVNN from two aspects: (i) view aggregation and (ii) training loss. For view aggregation, View-GCN and View-GCN++~\cite{wei2022learning} were proposed by treating the view CNN features as the graph nodes and applied graph convolution and view-sampling for view feature aggregation. SeqViews2SeqLabels~\cite{han2018seqviews2seqlabels} exploited RNN with attention to aggregate visual features by treating them as a sequence with temporal correlation.
In~\cite{lin2022multi}, View Attention Module (VAM) and Instance Attention Module (IAM) were designed to exploit view-relevant discriminative information and the instance-relevant correlative information for view feature aggregation. Other than CNN, Multi-View Vision Transformer (MVT)~\cite{Chen2021MVTMV} proposed to use vision transformer for view-associated feature extraction and aggregated them via a self-attention mechanism. Regarding training loss, metric learning based losses were widely explored, such as triplet-center loss~\cite{he2018triplet}, angular triplet-center loss~\cite{li2019angular}, triplet-center loss with adaptive margin~\cite{he2020improved} and cosine-distance-based triplet-center loss together with ArcFace loss~\cite{lin2022multi}.

Though the aforementioned methods have produced great performance in 3D shape recognition, they require a large number of training samples. However, collecting and annotating large-scale 3D data is laborious for industry applications. Hence, in this paper, we aim to address the more challenging task of multi-view based 3D shape recognition without training (zero-shot scenario) or training with a very limited number of data (few-shot scenario).

\begin{figure*}[!t]
  \centering
  % \vspace{8mm}
  \includegraphics[scale=0.5]{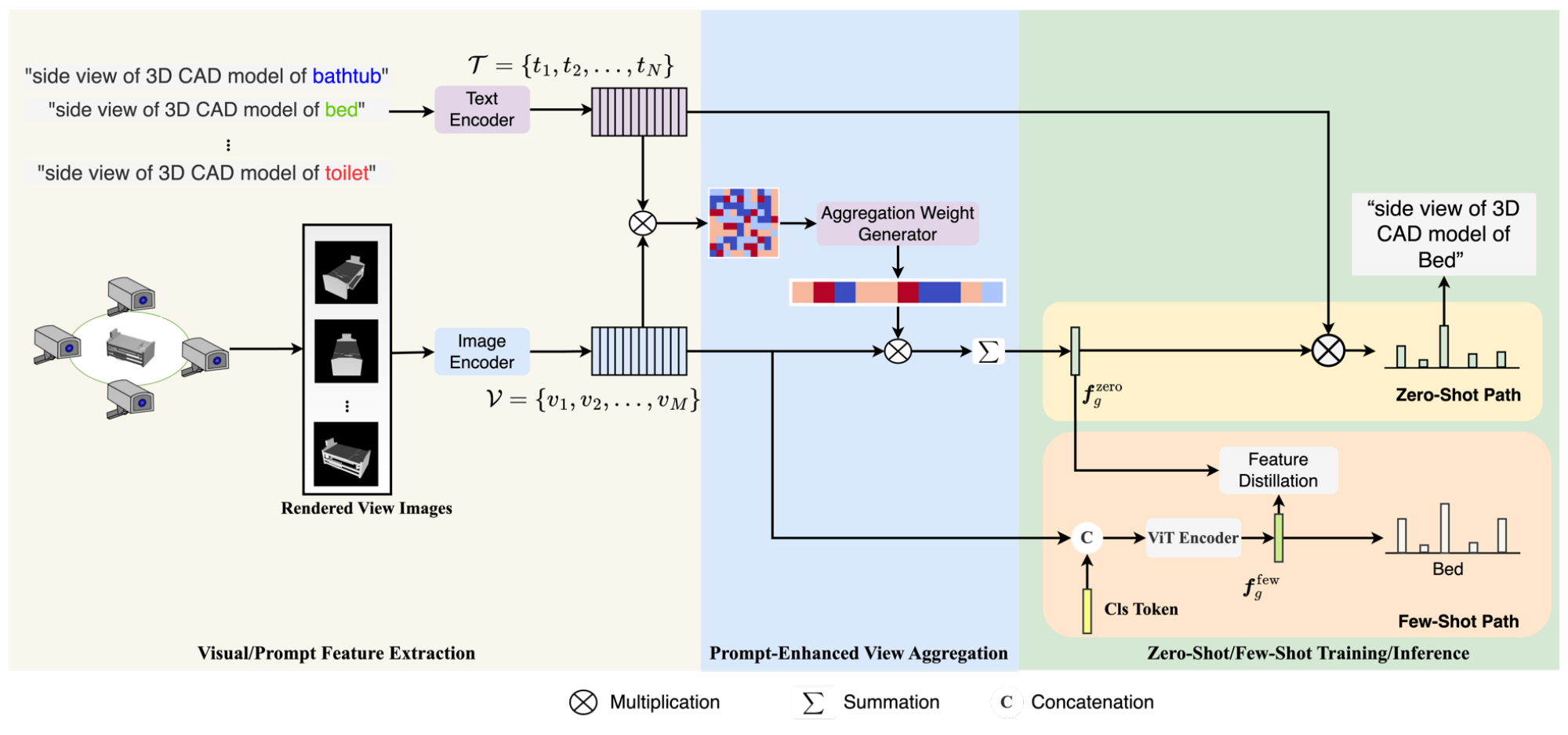}
  \caption{The overall architecture of the proposed PEVA-Net}
  \label{fig:Overall_Architecture}
\end{figure*}

\subsection{Zero/Few-Shot 3D Shape Recognition}
Recently, large-vision language models (LVLM), e.g., CLIP~\cite{radford2021learning}, have impressively promote the performance of 2D visual analysis under zero/few-shot scenarios for various applications, such as image classification~\cite{radford2021learning}, anomaly detection~\cite{jeong2023winclip} and video understanding~\cite{Xu2021VideoCLIPCP}. Motivated by these works, a growing number of works have emerged to investigate how to adapt CLIP onto 3D vision to address zero/few-shot 3D shape recognition. The relevant works can be grouped into two types, namely, pre-training based methods and pre-training-free methods.

Pre-training based methods focus on adapting the multi-modality contrastive learning proposed in CLIP~\cite{radford2021learning} onto the 3D multi-modality contrastive learning on large-scale 3D shape data encoded with advanced 3D deep learning network like Point Transformer\cite{zhao2021point}, CrossPoint~\cite{afham2022crosspoint}, Point-MAE~\cite{pang2022masked} and Point-BERT~\cite{yu2022point}. For example, CLIP2Point~\cite{huang2023clip2point} proposed to exploit a depth map encoder to replace the CLIP visual encoder in the contrastive learning paradigm and thereby to effectively transfer CLIP knowledge to 3D vision. CG3D~\cite{Hegde2023CLIPG3} introduced a 3D encoder (e.g., Point Transformer~\cite{zhao2021point} or Point-BERT~\cite{yu2022point}) into the original CLIP framework and pre-trained the 3D encoder together with the CLIP's text and image encoder in a triple contrastive learning manner. Similarly, ULIP~\cite{xue2023ulip} was proposed to learn a unified representations of image, text and point cloud via pre-training from these three modalities. It significantly improved the performance of 3D shape recognition for multiple 3D point cloud encoders, including PointNet++~\cite{QiPointNet++}, Point-Bert~\cite{yu2022point} and PointMLP~\cite{ma2022rethinking}. ULIP-2~\cite{xue2023ulip} further improved ULIP's language modality's scalability and comprehensiveness by leveraging large multi-modal models to generate the holistic language prompts. OpenShape~\cite{liu2024openshape} followed the ULIP framework and focused on scaling up representation learning via increasing data scale, enhancing the text quality, scaling up 3D backbones and resampling data. ReCon~\cite{qi2023contrast} exploited both the contrastive and generative pre-training paradigms via ensemble distillation to improve 3D representation learning. Though impressive performance has been produced by the pre-training based methods, they require large-scale datasets for pre-training and therefore are computationally expensive.

Pre-training-free methods exploited the multiple 2D view images as the 3D shape representation and directly leverage the CLIP model for zero/few-shot 3D recognition. PointCLIP~\cite{zhang2022pointclip} projected a point cloud onto multiple depth images and aggregate the CLIP visual features associated with each depth image to generate a descriptor for zero/few-shot 3D shape recognition. PointCLIP V2~\cite{zhu2023pointclip} further improved PointCLIP via prompting CLIP with a shape projection module to enhance the depth image generation and prompting the GPT model to generate the better prompts for 3D shape description. DiffCLIP~\cite{shen2024diffclip} proposed a style transfer process based on stable diffusion~\cite{rombach2022high} and ControlNet~\cite{zhang2023adding} to map the depth images into photorealistic 2D RGB images which are more suitable as the inputs for CLIP visual encoders. Besides depth images, the recently published Multi-View CLIP (MV-CLIP)~\cite{song2023mv} integrated CLIP into the MVCNN~\cite{su2015multi} paradigm by selecting a subset of the rendered 2D view images of the 3D shape based on entropy of the logit to generate a descriptor for zero-shot 3D shape recognition.

Our PEVA-Net belongs to the pre-training-free methods as no pre-training is required. Similar to MV-CLIP~\cite{song2023mv}, we also adopt rendered 2D view images as the 3D object representation. For zero-shot scenario, compared with MV-CLIP which only selected a subset of views, our PEVA-Net does not drop any views but provide a weighted aggregation scheme which leverages the prompt information to enhance the view aggregation. For few-shot scenario, compared with other pre-training-free methods~\cite{zhang2022pointclip,huang2023clip2point} which treated zero-shot and few-shot 3D shape recognition as isolated tasks, our PEVA-Net proposes a self-distillation scheme to exploit the zero-shot descriptor to guide the training of the few-shot descriptor via feature distillation and thereby to significantly improve the few-shot learning efficacy.

\section{Proposed Method}
\subsection{Overview}
As shown in Fig.~\ref{fig:Overall_Architecture}, the proposed PEVA-Net consists of three major stages in sequence, namely, Visual/Prompt Feature Extraction, Prompt-Enhanced View Aggregation and Zero/Few-Shot Training/Inference. The details of each stage are introduced in the following subsections.

\subsection{Visual and Prompt Feature Extraction}
As shown in the first stage of Fig.~\ref{fig:Overall_Architecture}, we exploit the CLIP text encoder and image encoder as the backbone networks to extract the visual and the prompt features, respectively. Given a 3D shape $\mathcal{O}$ with $M$ rendered view images denoted as $\mathcal{I} = \{\bm{I}_1, \bm{I}_2, \ldots, \bm{I}_M\}$, the view images are fed through the CLIP image encoder network to obtain the high-dimensional feature representations $\mathcal{V} = \{\bm{v}_1, \bm{v}_2, \ldots \bm{v}_M\}$, where
\begin{equation}
    \bm{v}_i = {\rm{Image\ Encoder}}(\bm{I}_i) \quad i=1, 2, \ldots, M.
\end{equation}

\noindent For text feature extraction, we leverage all the $N$ categorical labels $\mathcal{C} = \{c_1, c_2, \ldots, c_N\}$ to build up a prompt pool $\mathcal{P} = \{\bm{p}_1, \bm{p}_2, \ldots, \bm{p}_N\}$ with the prompt format as $\bm{p}_j = $ ``side view of 3D CAD model of $c_j$''. These prompts are then fed through the CLIP text encoder network to generate the prompt-associated feature representations $\mathcal{T} = \{\bm{t}_1, \bm{t}_2, \ldots, \bm{t}_N\}$, where
\begin{equation}
    \bm{t}_j = {\rm{Text\ Encoder}}(\bm{p}_j)\quad j=1, 2, \ldots, N.
\end{equation}

% ``side view of 3D CAD model of $\{{\rm{CLASS}}\}$, where ${\rm{CLASS}}$ is replaced by the categorical label of the 3D object. Formally, given all the categorical labels $\mathcal{C} = \{c_1, c_2, \ldots, c_N\}$, the prompt pool is generated as $\mathcal{P} = \{\bm{p}_1, \bm{p}_2, \ldots, \bm{p}_N\}$, where $\bm{p}_j$ is ``side view of 3D CAD model of $c_j$''.

\subsection{Prompt-Enhanced View Aggregation}

With the extracted visual and prompt features, we propose to exploit the prompt features to aggregate the view-associated visual features into a discriminative descriptor for zero-shot and few-shot 3D shape recognition. Specifically, given object $\mathcal{O}$, the visual features $\mathcal{V} = \{\bm{v}_1, \bm{v}_2, \ldots \bm{v}_M\}$ and the prompt features $\mathcal{T} = \{\bm{t}_1, \bm{t}_2, \ldots, \bm{t}_N\}$ for the prompt pool are generated. Then, the visual-prompt similarity matrix $\bm{S}$ can be generated with its entry $\bm{S}_{ij}$ being calculated as the inner product of $\bm{t}_i$ and $\bm{v}_j$:
\begin{equation}
    \bm{S}_{ij} =  \bm{t}_i^T\bm{v}_j.
\end{equation}
\noindent The column vectors $\bm{s}_1, \bm{s}_2, \ldots \bm{s}_M$ of $\bm{S}$ indicate the similarity between each visual feature and the prompt features. To quantify the discriminative score for each visual feature, we define the weight $\bm{A} = \{\alpha_1, \alpha_2, \ldots, \alpha_M\}$ for $\mathcal{V} = \{\bm{v}_1, \bm{v}_2, \ldots \bm{v}_M\}$ as:
\begin{equation}
    \alpha_i = \bm{s}_i[{\rm{argmax}}(\bm{s}_i)] - \bar{\bm{s}_i},
\end{equation}
where $\bm{s}_i[{\rm{argmax}}(\bm{s}_i)]$ selects the maximum entry value of $\bm{s}_i$ and $\bar{\bm{s}_i}$ denotes the average value of all the entries of $\bm{s}_i$, i.e., $\bar{\bm{s}_i} = \frac{1}{N}\sum_{j=1}^{N}{\bm{s}_i[j]}$. The higher value of the discriminative score indicates the better visual features in terms of the confidence for zero-shot recognition under the CLIP rule. To aggregate the view-associated visual features into a global descriptor for zero-shot recognition, the aggregation weights $\bm{w} = \{w_1, w_2, \ldots, w_M\}$ can be calculated via a softmax operation upon the entries of $\bm{A}$:
\begin{equation}
    w_i = \frac{e^{\alpha_i}}{\sum_{j=1}^{M} e^{\alpha_j}}.
\end{equation}
Then, the aggregated feature for zero-shot recognition can be calculated by the weighted sum of visual features:
\begin{equation}
    \bm{f}^{\rm{zero}} = \sum_{i=1}^{M} w_i \bm{v}_i
\end{equation}

\subsection{Zero-Shot 3D Shape Recognition}
\label{sec:zero-shot}
With the aggregated descriptor $\bm{f}^{\rm{zero}}$ for 3D shape $\mathcal{O}$, zero-shot recognition of $\mathcal{O}$ can be conducted using the CLIP inference rule by computing the inner product between the prompt features $\mathcal{T} = \{\bm{t}_1, \bm{t}_2, \ldots, \bm{t}_N\}$ and $\bm{f}^{\rm{zero}}$. Specifically, the logits for zero-shot inference $\bm{L}^{\rm{zero}} = \{l_1^{\rm{zero}}, l_2^{\rm{zero}}, \ldots, l_N^{\rm{zero}}\}$ can be computed as:
\begin{equation}
    l_j^{\rm{zero}} = \bm{t}_{j}^{T}\bm{f}^{\rm{zero}} \quad j=1, 2, \ldots, N.
    \label{zero_inference}
\end{equation}

\noindent The predicted label of the object will be the one which can produce the largest logit.

\subsection{Few-Shot 3D Shape Recognition}
The proposed PEVA-Net is also capable of few-shot multi-view 3D shape recognition. Here, we follow the design of few-shot setting in PointCLIP~\cite{zhang2022pointclip} where the model is trained using the training data under ``K-Shot'' setting and tested using the entire testing data. ``K-Shot'' setting refers to the case that for each category, only K samples are selected to construct the training set.

% In general, we leverage the highly generalized CLIP visual features for each view image and the proposed prompt-enhanced aggregated visual features to learn a discriminative representation of 3D shape under the few-shot setting.

\begin{figure}[htbp]
  \centering
  % \vspace{8mm}
  \includegraphics[scale=0.45]{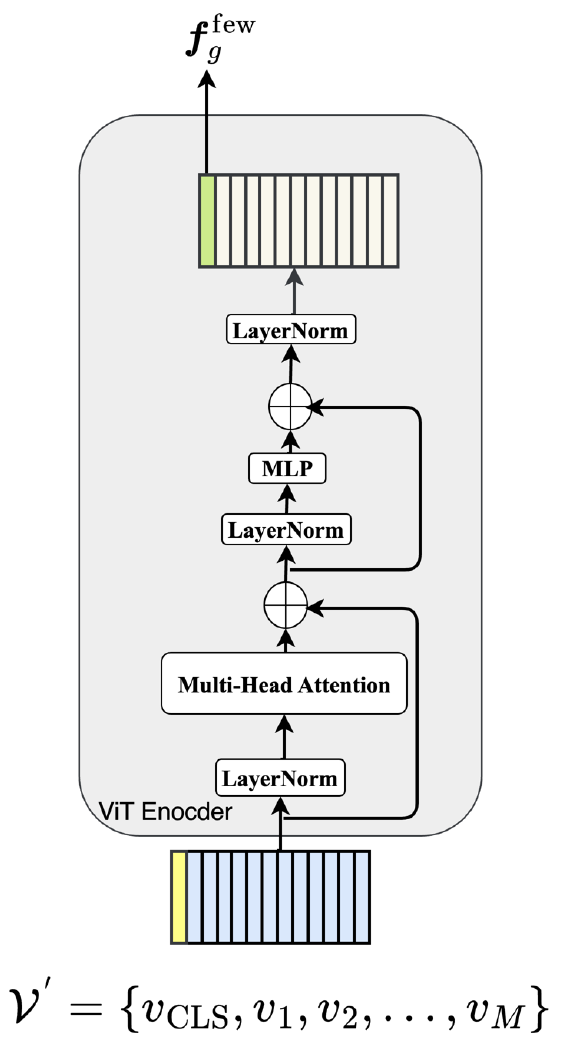}
  \caption{The architecture of ViT Encoder.}
  \label{fig:ViT_Encoder}
\end{figure}

As shown in Fig.\ref{fig:Overall_Architecture}, with the 3D shape $\mathcal{O}$ represented by M visual features $\mathcal{V} = \{\bm{v}_1, \bm{v}_2, \ldots \bm{v}_M\}$, we first prepend a CLS token onto $\mathcal{V}$ to generate $\mathcal{V}^{'} = \{\bm{v}_{\rm{CLS}}, \bm{v}_1, \bm{v}_2, \ldots \bm{v}_M\}$. Then $\mathcal{V}^{'}$ is fed through a Vision Transformer (ViT) Encoder where visual features are interacted with each other via the self-attention mechanism introduced in ~\cite{dosovitskiy2021an} and all the information is aggregated into the CLS token associated feature vector. We apply the CLS token associated feature vector $\bm{f}^{\rm{few}}$ to conduct few-shot multi-view 3D shape recognition. Formally, $\bm{f}^{\rm{few}}$ is obtained via:
\begin{equation}
    \bm{f}^{\rm{few}} = {\rm{Encoder}}(\mathcal{V}^{'};\theta_{\rm{ViT}}),
\end{equation}
where $\theta_{\rm{ViT}}$ denotes that the trainable parameters in the ViT Encoder. Following~\cite{dosovitskiy2021an} The ViT Encoder consists of the LayerNorm module, Multi-Head Attention module and MLPs in cascade. The aggregation process is illustrated in Fig.\ref{fig:ViT_Encoder}. The detailed formulation of the encoder based visual feature aggregation is specified in the supplementary material.

To enhance the few-shot training efficacy, we propose a self-distillation scheme to exploit the zero-shot descriptor $\bm{f}^{\rm{zero}}$ as the guiding signal to tune the few-shoe descriptor $\bm{f}^{\rm{few}}$ via a feature distillation process. Specifically, the few-shot training process of PEVA-Net involves two losses: (i) the classification loss $\mathcal{L}_{\rm{cls}}$ between the prompt features $\mathcal{T} = \{\bm{t}_1, \bm{t}_2, \ldots, \bm{t}_N\}$ and $\bm{f}^{\rm{few}}$; and (ii) the feature distillation loss $\mathcal{L}_{\rm{fd}}$ between $\bm{f}^{\rm{zero}}$ and $\bm{f}^{\rm{few}}$. The overall loss $\mathcal{L}_{\rm{few}}$ for few-shot training could be defined as
\begin{equation}
    \mathcal{L}_{\rm{few}} = \mathcal{L}_{\rm{cls}} + \mathcal{L}_{\rm{fd}}.
\end{equation}
\noindent During inference, the same inference rule for zero-shot setting in Eq.~\ref{zero_inference} can be applied by replacing $\bm{f}^{\rm{zero}}$ by $\bm{f}^{\rm{few}}$. Here, we provide the detailed formulations of $\mathcal{L}_{\rm{cls}}$ and $\mathcal{L}_{\rm{fd}}$, respectively.

\subsubsection{\textbf{Classification Loss}}

We define the classification loss as the cross-entropy loss with the logits calculated via the inner product between the prompt features $\mathcal{T}$ and $\bm{f}^{\rm{few}}$. Similar to the zero-shot inference in Eq. \ref{zero_inference}, the logits for few-shot training $\bm{L}^{\rm{few}} = \{l_1^{\rm{few}}, l_2^{\rm{few}}, \ldots, l_N^{\rm{few}}\}$ can be computed as:
\begin{equation}
    l_j^{\rm{few}} = \bm{t}_{j}^{T}\bm{f}^{\rm{few}} \quad j=1, 2, \ldots, N.
\end{equation}
Then, the softmax operation is applied on every entry of  $l_j^{\rm{few}}$ to generate the predicted probability $\hat{l}_j^{\rm{few}}$, i.e., $\hat{l}_j^{\rm{few}} = \frac{e^{l_j^{\rm{few}}}}{\sum_{n=1}^{N} e^{l_n^{\rm{few}}}}$. Finally, with $\hat{\bm{L}}^{\rm{few}} = \{\hat{l}_1^{\rm{few}}, \hat{l}_2^{\rm{few}}, \ldots, \hat{l}_N^{\rm{few}}\}$, the classification loss is defined as the cross-entropy loss with respect to $\hat{\bm{L}}^{\rm{few}}$ and the groundtruth label $y_j$ as:
\begin{equation}
    \mathcal{L}_{\rm{cls}} = -\mathds{1}[k = y_j]log(\hat{l}_k^{\rm{few}}),
\end{equation}
where $\mathds{1}(\cdot)$ denotes the indication function. The classification loss is designed based on the CLIP inference rule to facilitate the encoder to generate the aggregated visual features to align with the semantic information provided by the prompt features. It is adopted as the main loss function to finetune the PEVA-Net under the few-shot scenario.

% Table generated by Excel2LaTeX from sheet 'vs_mv_clip'
\begin{table*}[!t]
  \centering
  \caption{Zero-Shot Recognition Performance (in \%) on ModelNet40 and ModelNet10}
  \scalebox{0.9}{
    \begin{tabular}{cccccc}
    \toprule
    \multirow{2}[4]{*}{Method} & \multirow{2}[4]{*}{CLIP version} & \multicolumn{2}{c}{\multirow{2}[4]{*}{Pre-Training Source}} & \multicolumn{2}{c}{Accuracy} \\
\cmidrule{5-6}          &       & \multicolumn{2}{c}{} & ModelNet40 & ModelNet10 \\
    \midrule
    CG3D~\cite{Hegde2023CLIPG3} + Point Transformer~\cite{zhao2021point} & SLIP  & \multicolumn{2}{c}{ShapeNet} & 50.60 & -  \\
    ULIP~\cite{xue2023ulip} + Point-BERT~\cite{yu2022point} & SLIP  & \multicolumn{2}{c}{ShapeNet} & 60.40 & -  \\
    ULIP-2~\cite{xue2023ulip} + Point-BERT~\cite{yu2022point} & SLIP  & \multicolumn{2}{c}{ShapeNet} & 66.40 & - \\
    ULIP-2~\cite{xue2023ulip} + Point-BERT~\cite{yu2022point} & SLIP  & \multicolumn{2}{c}{Objaverse} & 74.00 & -  \\
    OpenShape~\cite{liu2024openshape} + Point-BERT~\cite{yu2022point} & OpenCLIP & \multicolumn{2}{c}{ShapeNet} & 72.90 & - \\
    CLIP2Point~\cite{huang2023clip2point} & CLIP  & \multicolumn{2}{c}{ShapeNet} & 49.38 & 66.63 \\
    Recon~\cite{qi2023contrast} & CLIP  & \multicolumn{2}{c}{ShapeNet} & 61.70 & 75.60 \\
    PointCLIP~\cite{zhang2022pointclip} & CLIP  & \multicolumn{2}{c}{-} & 20.18 & 30.23 \\
    PointCLIP V2~\cite{zhu2023pointclip} & CLIP  & \multicolumn{2}{c}{-} & 64.22 & 73.13 \\
    DiffCLIP~\cite{shen2024diffclip} & CLIP  & \multicolumn{2}{c}{-} & 49.70 & 80.60 \\
    MV-CLIP~\cite{song2023mv} & CLIP  & \multicolumn{2}{c}{-} & 65.92 & 77.53 \\
    MV-CLIP~\cite{song2023mv} & OpenCLIP & \multicolumn{2}{c}{-} & 84.44 & 91.51 \\
    \midrule
    Our PEVA-Net & CLIP  & \multicolumn{2}{c}{-} & \textbf{66.12} & \textbf{82.26} \\
    Our PEVA-Net & OpenCLIP & \multicolumn{2}{c}{-} & \textbf{84.48} & \textbf{93.50} \\
    \bottomrule
    \end{tabular}%
  \label{tab:zero-shot-modelnet}%
  }
\end{table*}%

\subsubsection{\textbf{Feature Distillation Loss}}

To further enhance the efficacy of few-shot learning and alleviate overfitting, we propose a self-distillation scheme by facilitating feature distillation between the prompt-enhanced aggregated feature $\bm{f}^{\rm{zero}}$ and the encoder aggregated feature $\bm{f}^{\rm{few}}$. Specifically, the feature distillation loss $\mathcal{L}_{\rm{fd}}$ is defined as:
\begin{equation}
    \mathcal{L}_{\rm{fd}} = ||\bm{f}^{\rm{few}} - \bm{f}^{\rm{zero}}||^2.
\end{equation}

\noindent Since $\bm{f}^{\rm{zero}}$ is effective for zero-shot recognition, the feature distillation loss servers as a regularization term to guide the aggregated feature $\bm{f}^{\rm{few}}$ to converge to a solution which is close to $\bm{f}^{\rm{zero}}$ and finetuned using a limited number of training data.

% The overall loss $\mathcal{L}_{\rm{few}}$ for few-shot training could be defined as
% \begin{equation}
%     \mathcal{L}_{\rm{few}} = \mathcal{L}_{\rm{cls}} + \mathcal{L}_{\rm{fd}}.
% \end{equation}

\section{Experiments}
In this section, the extensive experiments are conducted to show the superiority of the proposed PEVA-Net on zero-shot and few-shot recognition of 3D shapes. We also conduct the ablation studies and analysis to verify the effectiveness of the proposed techniques.

\subsection{Datasets and Evaluation Metrics}
For zero-shot 3D shape recognition, we adopt ModelNet10, ModelNet40 and ShapeNetCore 55 for the experiments following the prior work~\cite{song2023mv}. These datasets consists of 3D shapes from common categories in the format of 3D CAD mesh models. ModelNet10 and ModelNet40 are subsets of the Princeton ModelNet data set constructed with 3D CAD objects. Specifically, ModelNet10 contains 4,899 3D shapes from 10 categories, with 908 objects in the testing set while ModelNet40 contains 12,311 3D shapes from 40 categories, with 2468 objects for testing. ShapeNetCore 55 contains 51,162 3D shapes from 55 categories with 10,625 objects for testing. For few-shot 3D shape recognition, following the work of CLIP2Point~\cite{huang2023clip2point}, we adopt ModelNet40 with the shot number K set as 16 to build up the few-shot training set. In comparison with PointCLIP~\cite{zhang2022pointclip}, the shot number is set as $\{8, 16, 32, 64, 128\}$. The testing set is the entire testing set of ModelNet40. For all the experiments, the classification accuracy on the testing set is adopted as the evaluation metric. The inference rule follows Eq.~\ref{zero_inference}, similar to the work of CLIP~\cite{radford2021learning}.

% of PointCLIP~\cite{zhang2022pointclip}, we adopt ModelNet40 with the shot number K set as 8, 16, 32, 64 and 128 to build up the few-shot training set, respectively.

\subsection{Implementation Details}
All the experiments are run on four NVIDIA Tesla V100 GPUs and implemented using the Pytorch framework. For zero-shot experiment, to generate the zero-shot descriptor using the proposed prompt-enhanced view aggregation, we create the prompt with the format of ``A side view of 3D CAD model of \{CLASS\}'', where CLASS can be replaced by the object category, such as ``bathtub'' or ``bed''. For few-shot experiment, in the encoder, we did not exploit any position embedding since the inputs are view-associated visual features. For multi-head attention module, the projection dimension for the query, key and value embeddings are set as 1024. The number of of head is set as 4. The MLP output dimension is set as 512. During training, Adam optimizer is adopted with the learning rate set as 0.001, the momentum is set as 0.9, the weight decay set as 0.0001. The total epoch number is set as 50. Only the parameters in the encoder are tuned while the remaining parameters of the PEVA-Net are freezed without tuning. The feature distillation from the zero-shot descriptor to the few-shot descriptor is only conducted during the training. We consider two versions of CLIP for our PEVA-Net, namely, the vanilla CLIP~\cite{radford2021learning} and OpenCLIP~\cite{cherti2023reproducible}.

\subsection{Zero-Shot 3D Shape Recognition}
Table~\ref{tab:zero-shot-modelnet} shows the zero-shot classification accuracy produced by the competing methods on ModelNet40 and ModelNet10, respectively. From the table, it is observed that our PEVA-Net based on OpenCLIP can achieve the best classification performance of 84.48\% and 93.50\% on ModelNet40 and ModelNet10, respectively. It is also noted that such superior performance does not rely on any pre-training dataset of 3D shape recognition, such as ShapeNet or Objaverse. Specifically, compared with the best competing method with pre-training dataset, i.e., ULIP-2~\cite{xue2023ulip} + Point-BERT~\cite{yu2022point} pretrained using Objaverse, our PEVA-Net can produce the performance gain of 10.48\% on ModelNet40 dataset. Compared with OpenShape~\cite{liu2024openshape} + Point-BERT which is pretrained on ShapeNet and also adopts OpenCLIP, our PEVA-Net achieves the performance gain of 11.58\%. Compared with those state-of-the-art methods based on the vanilla CLIP and do not exploit any pretraining dataset, the proposed PEVA-Net based on the vanilla CLIP also achieves the best classification performance.

\begin{table}[htbp]
  \centering
  \caption{Zero-Shot Recognition Performance (in \%) on ShapeNetCore 55}
  \scalebox{1.0}{
    \begin{tabular}{ccc}
    \toprule
    Method & CLIP version & Accuracy \\
    \midrule
    MV-CLIP~\cite{song2023mv} & CLIP  & 61.70 \\
    Our PEVA-Net & CLIP  & \textbf{66.49} \\
    \midrule
    MV-CLIP~\cite{song2023mv} & OpenCLIP & 66.17 \\
    Our PEVA-Net & OpenCLIP & \textbf{74.65} \\
    \bottomrule
    \end{tabular}%
  \label{tab:zero-shot-shapenet}%
  }
\end{table}%

In comparison with the recently published MV-CLIP~\cite{song2023mv}, under both CLIP versions, our PEVA-Net can outperform MV-CLIP on both ModelNet datasets. Table~\ref{tab:zero-shot-shapenet} shows the zero-shot recognition performance on ShapeNetCore 55 dataset. From the table, it is observed that our PEVA-Net could significantly outperform MV-CLIP by the margin of 4.79\% and 8.16\% under the vanilla CLIP and OpenCLIP settings, respectively. In MV-CLIP, only a subset of views are selected based on entropy of the logit for zero-shot 3D shape classification. Therefore, the information within the views with higher entropy values is dropped, which may lead to the deteriorated recognition performance. In contrary, our PEVA-Net does not drop any views but provide a weighted aggregation scheme leveraging the prompt information to enhance the view aggregation process to boost the recognition performance.

% The experimental results show the superiority of the proposed prompt-enhanced view aggregation.

\subsection{Few-Shot 3D Shape Recognition}
In this section, we compare our PEVA-Net with several competing methods under the few-shot 3D shape recognition scenario. Specifically, we compare our method with two self-supervised pre-training methods (CrossPoint~\cite{afham2022crosspoint} and Point-MAE~\cite{pang2022masked}) and two supervised methods (PointCLIP~\cite{zhang2022pointclip} and CLIP2Point~\cite{zhang2022pointclip}). Here, following the prior work of CLIP2Point, CrossPoint adopts DGCNN as the backbone network and Point-MAE adopts a 12-layer transformer encoder. PointCLIP and CLIP2Point adopt ResNet101 and ViT-B/32 as the backbone networks since these backbone settings could produce the best few-shot recognition performance. Table~\ref{tab:16_shot_performance} presents the 16-shot recognition accuracy of 3D shapes on ModelNet40. From the table, it is observed that our PEVA-Net could produce the best performance (90.64\%) among all the compared methods. Without any pre-training, the proposed PEVA-Net significantly outperforms the self-supervised pre-training methods CrossPoint and Point-MAE by 6.16\% and 6.44\%, respectively. Compared with the supervised CLIP based methods, our method outperforms PointCLIP and CLIP2Point by 3.44\% and 3.18\%, respectively. In addition, our PEVA-Net without pre-training can still slightly outperform the performance produced by CLIP2Point pretrained on ShapeNet. These experimental observations verify the superiority of the proposed few-shot learning scheme for 3D shape recognition involving the transformer aggregation together with the effective feature distillation from the zero-shot descriptor.

% Table generated by Excel2LaTeX from sheet 'vs_mv_clip'
\begin{table}[htbp]
  \centering
  \caption{16-Shot Recognition Performance (in \%) on ModelNet40}
  \scalebox{0.9}{
    \begin{tabular}{ccc}
    \toprule
    Method & w/o Pre. & w/Pre. \\
    \midrule
    CrossPoint~\cite{afham2022crosspoint} & 81.56 & 84.48 \\
    Point-MAE~\cite{pang2022masked} & 79.70  & 84.20 \\
    PointCLIP~\cite{zhang2022pointclip} & 87.20  & -  \\
    CLIP2Point~\cite{huang2023clip2point} & 87.46 & 89.79 \\
    \midrule
    Our PEVA-Net & \textbf{90.64} & - \\
    \bottomrule
    \end{tabular}%
    }
  \label{tab:16_shot_performance}%
\end{table}%
\begin{figure}[htbp]
  \centering
  % \vspace{8mm}
  \includegraphics[scale=0.6]{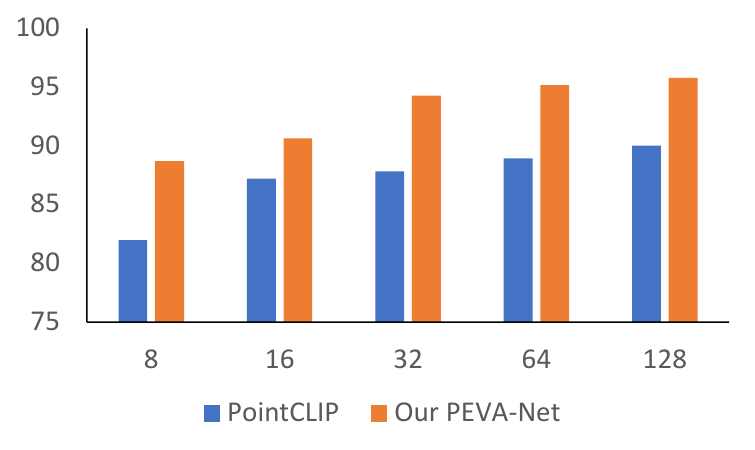}
  \caption{Few-Shot Recognition Performance on ModelNet40 Under Different Number of Shots.}
  \label{fig:few_shot_with_pointclip}
\end{figure}

We further conduct an experiment to show the performance of PEVA-Net under varied shot number settings. Fig.~\ref{fig:few_shot_with_pointclip} shows the performance of our PEVA-Net and PointCLIP~\cite{zhang2022pointclip} with respect to different number of shots on ModelNet40. From the figure, it is demonstrated that our PEVA-Net can produce the better performance than PointCLIP across all the shot numbers. Particularly, our method achieves 95.78\% classification accuracy under 128-shot setting, which is comparable to the state-of-the-art performance reached using the full training set for model learning.

% Table generated by Excel2LaTeX from sheet 'vs_mv_clip'
% \begin{table}[htbp]
%   \centering
%   \caption{Few-Shot Recognition Performance on ModelNet40}
%     \begin{tabular}{|cccccc|}
%     \toprule
%     \# of Shots & 8     & 16    & 32    & 64    & 128 \\
%     \midrule
%     \midrule
%     PointCLIP & 81.96 & 87.2  & 87.83 & 88.95 & 90.02 \\
%     Our PEVA-Net & 88.69 & 90.64 & 94.28 & 95.18 & 95.78 \\
%     \bottomrule
%     \end{tabular}%
%   \label{tab:addlabel}%
% \end{table}%

\subsection{Discussion}\label{Discussion}

In this section, we conduct several ablation experiments to analyze the proposed PEVA-Net.

\subsubsection{\textbf{On Prompt-Enhanced View Aggregation}}
\label{subsec:ablation_on_PEVA}
In this subsection, we conduct an ablation experiment to show the effectiveness of the proposed scheme of Prompt-Enhanced View Aggregation (PEVA). We consider two ablation models: (i) the model which aggregates visual features via average pooling; (ii) the model which aggregates visual features via PEVA. Table~\ref{tab:ablation_on_PEVA} presents the zero-shot recognition accuracy for the ablation models with different view aggregation schemes on three benchmarking datasets. It is shown that the model with PEVA can outperform that with average pooling on all the datasets, thereby showing the effectiveness of prompt enhancement on visual feature aggregation.

% Table generated by Excel2LaTeX from sheet 'vs_mv_clip'
\begin{table}[htbp]
  \centering
  \caption{Zero-Shot Performance (in \%) for the Models with Different View Aggregation Scheme}
  \scalebox{0.95}{
    \begin{tabular}{cccc}
    \toprule
    Method & ModelNet40 & ModelNet10 & ShapeNetCore 55 \\
    \midrule
    Average Pooling & 61.83 & 79.84 & 65.45 \\
    PEVA (Ours) & \textbf{66.12} & \textbf{82.26} & \textbf{66.49} \\
    \bottomrule
    \end{tabular}%
    }
  \label{tab:ablation_on_PEVA}%
\end{table}%

\subsubsection{\textbf{On Prompt Design}}
In this subsection, we investigate the relationship between the prompt design and the performance gain produced by the proposed PEVA for zero-shot 3D shape recognition. Specifically, we test the zero-shot recognition performance on ModelNet40 for the models with four prompt designs as:
\begin{itemize}
    \item A photo of \{CLASS\}
    \item A project view of \{CLASS\}
    \item A project view of 3D CAD model of \{CLASS\}
    \item A side view of 3D CAD model of \{CLASS\}
\end{itemize}

\begin{figure}[htbp]
  \centering
  % \vspace{8mm}
  \includegraphics[scale=0.57]{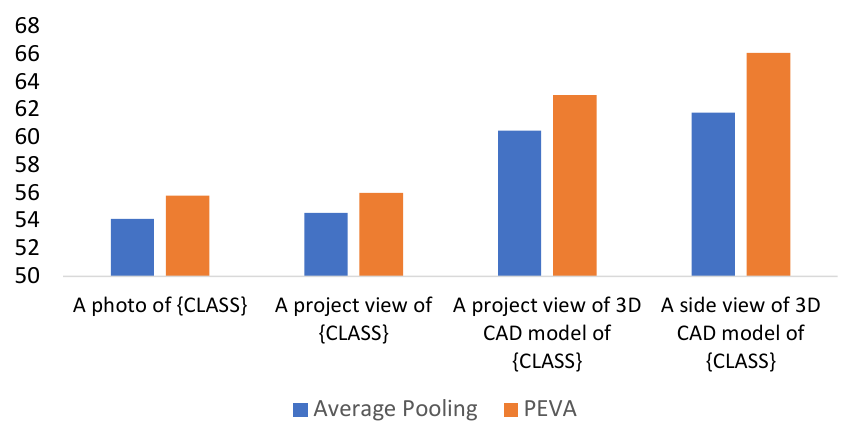}
  \caption{Zero-Shot Recognition Performance (in $\%$) for the Models with Different Prompt Designs.}
  \label{fig:ablation_on_prompts}
\end{figure}

Fig.~\ref{fig:ablation_on_prompts} shows the zero-shot recognition performance produced by the models with different prompt designs. We consider the same ablation models as in subsection~\ref{subsec:ablation_on_PEVA} with average pooling or the proposed PEVA as the aggregation operations. From the figure, it is first observed that for all the prompts, the model with PEVA can consistently perform better than the model with average pooling. Secondly, the performance gain would be significantly improved when a more 3D-shape-specific prompt is adopted, e.g., ``A side view of 3D CAD model of \{CLASS\}'', compared with the more generic description prompt like ``A photo of \{CLASS\}''. This observation sheds light on the potential that PEVA-Net could be tuned via prompt engineering to fit different domain-specific tasks without pre-training on large-scale domain-specific data.

\subsubsection{\textbf{On Backbone Network}}
In this subsection, we conduct the ablation experiments to verify the effectiveness of the proposed PEVA-Net across different image encoder backbone networks for CLIP. Here, we adopt the vanilla CLIP~\cite{radford2021learning} as the image and text encoders. Table~\ref{tab:ablation_on_backbone} shows the zero-shot recognition performance on ModelNet40 for the models with different backbone networks including RN50, RN101, ViT-B/32, ViT-L/16, ViT-L/14. It is clearly shown from the table that compared with the baseline method using average pooling, incorporating the proposed PEVA can consistently improve the performance across all the CNN and Vision Transformer based backbone networks.

% Table generated by Excel2LaTeX from sheet 'vs_mv_clip'
\begin{table}[htbp]
  \centering
  \caption{Zero-Shot Recognition Performance (in $\%$) for the Models with Different BackBone Networks}
  \scalebox{0.71}{
    \begin{tabular}{ccccc}
    \toprule
    \multirow{2}[2]{*}{Method} & \multirow{2}[2]{*}{BackBone Network} & \multirow{2}[2]{*}{ModelNet40} & \multirow{2}[2]{*}{ModelNet10} & \multirow{2}[2]{*}{ShapeNetCore 55} \\
          &       &       &       &  \\
    \midrule
    \multirow{5}[2]{*}{Average Pooling} & RN50  & 48.58 & 61.56 & 41.92 \\
          & RN101 & 52.63 & 70.70 & 50.08 \\
          & ViT-B/32 & 60.12 & 66.85 & 54.33 \\
          & ViT-L/16 & 55.38 & 72.13 & 58.19 \\
          & ViT-L/14 & 61.83 & 79.84 & 65.45 \\
    \midrule
    \multirow{5}[2]{*}{PEVA} & RN50  & \textbf{50.36} & \textbf{61.67} & \textbf{42.88} \\
          & RN101 & \textbf{54.25} & \textbf{72.13} & \textbf{52.13} \\
          & ViT-B/32 & \textbf{62.07} & \textbf{68.94} & \textbf{56.12} \\
          & ViT-L/16 & \textbf{57.29} & \textbf{74.00} & \textbf{59.89} \\
          & ViT-L/14 & \textbf{66.12} & \textbf{82.26} & \textbf{66.49} \\
    \bottomrule
    \end{tabular}%
    }
  \label{tab:ablation_on_backbone}%
\end{table}%

\subsubsection{\textbf{On Loss}}
In this subsection, we conduct the ablation experiments on ModelNet40 to analyze the function of the proposed self-distillation scheme. To this end, we test the performance of few-shot 3D shape recognition produced by our PEVA-Net trained using (i) only the classification loss $\mathcal{L}_{\rm{cls}}$ and (ii) the summation of the classification loss $\mathcal{L}_{\rm{cls}}$ and the feature distillation loss $\mathcal{L}_{\rm{fd}}$. From Table~\ref{tab:ablation_on_loss}, it is shown that training with distillation loss could consistently improve the few-shot classification performance across all the shot settings. Also, as illustrated in Fig.~\ref{fig:model_acc_visualization_and_tsne}(a), the distillation loss could facilitate the model to rapidly converge to the model with higher testing accuracy. These observations demonstrate that distillating knowledge from the zero-shot descriptor could effectively enhance the few-shot learning efficacy.

\begin{table}[htbp]
  \centering
  \caption{Few-Shot Recognition Performance (in \%) for the Ablation Models Trained using Different Losses}
  \scalebox{1.0}{
    \begin{tabular}{ccccccc}
    \toprule
    \multicolumn{2}{c}{\# of Shots} & 8     & 16    & 32    & 64    & 128 \\
     \midrule
    $\mathcal{L}_{\rm{cls}}$ & $\mathcal{L}_{\rm{fd}}$ &       &       &       &       &  \\

    \checkmark   &     & 87.84 & 90.19 & 93.76 & 94.65 & 95.46 \\
    \checkmark   & \checkmark   & \textbf{88.69} & \textbf{90.64} & \textbf{94.28} & \textbf{95.18} & \textbf{95.78} \\
    \bottomrule
    \end{tabular}%
    }
  \label{tab:ablation_on_loss}%
\end{table}%

% Interestingly, the highest performance gain is achieved for the 8-shot case where only 8 training samples are picked for training for each category.

% The reason could be the model training can benefit most from the regularization of feature distillation when only very limited training samples are provided.

% Table generated by Excel2LaTeX from sheet 'vs_mv_clip'

\subsubsection{\textbf{Visualization of View Images with Prompt Enhancement Weights}}

\begin{figure}[!t]
  \centering
  % \vspace{8mm}
  \includegraphics[scale=0.29]{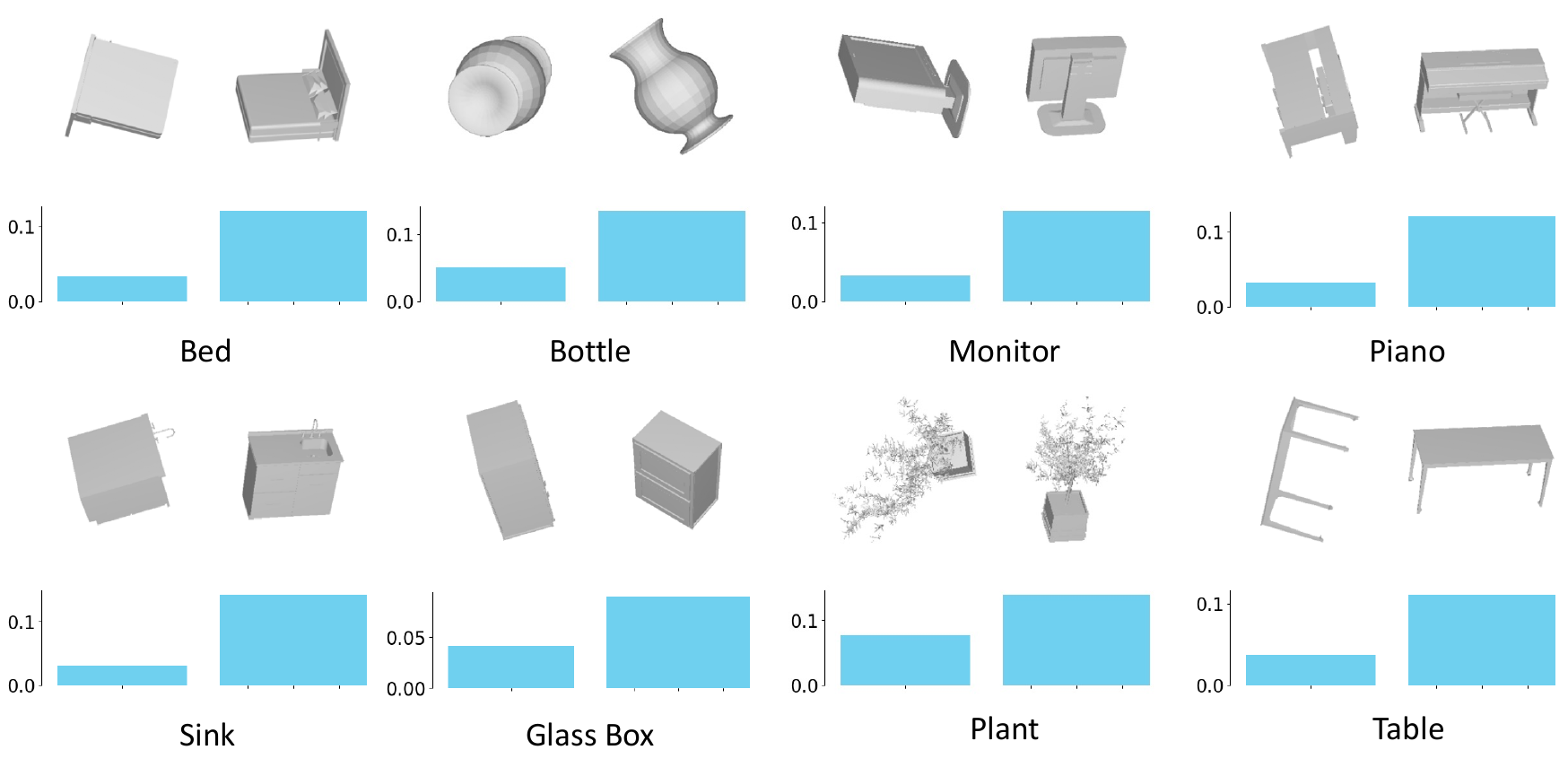}
  \caption{Examples of view images of 3D shapes with their discriminative scores. The examples are picked from the testing set of ModelNet40.}
  \label{fig:weight_view_visualization}
\end{figure}

In this subsection, we present a qualitative visualization of view images for some 3D shapes with respect to their prompt enhancement weights in Fig.~\ref{fig:weight_view_visualization}. For simplicity, for each 3D shape, we only present the view images associated with the lowest and the highest discriminative scores, respectively. From the figure, it is clearly observed that the proposed aggregation weights can effectively quantify the discriminative capability of the view images. For example, for a 3D shape of a bed, the lowest discriminative score is assigned to the view image containing only the bottom of the bed which is not informative while the highest score corresponds to a view image of an upright bed with the pillows and the backboard. The more detailed view-weight pair visualizations are provided in the supplementary material.

% \subsubsection{\textbf{Visualization of Learned Visual Embeddings}}
% In this subsection, we plot the 3D t-SNE embeddings for the testing samples in 10 out of 40 categories in ModelNet40 learned by our PEVA-Net under zero-shot and 16-shot scenarios, respectively. From the figure, it is observed that our PEVA-Net could generate the discriminative feature representations under the zero-shot scenario. With only a few training samples, the proposed PEVA-Net could efficiently learn to fit the domain-specific 3D shapes without overfitting.

% \begin{figure}[htbp]
%   \centering
%   % \vspace{8mm}
%   \includegraphics[scale=0.42]{t_sne_embeddings}
%   \caption{The 3D t-SNE embeddings for (i) zero-shot scenario and (ii) 16-shot scenario.}
%   \label{fig:t_sne_embeddings}
% \end{figure}

\section{Conclusion}
In this paper, we propose a Prompt-Enhanced View Aggregation Network (PEVA-Net) based on CLIP to simultaneously address zero/few-shot 3D shape recognition. Firstly, we leverage the prompt (text) information to enhance the aggregation process of the multi-view (visual) features of 3D shapes for effective zero-shot recognition. Secondly, for few-shot 3D shape recognition, we propose to leverage the zero-shot descriptor to guide the training of the few-shot descriptor via feature distillation to significantly improve the few-shot learning efficacy. The extensive experiments demonstrate that our PEVA-Net can achieve the state-of-the-art zero/few-shot recognition performance on multiple benchmarking datasets without any pre-training process. One future direction to further improve our PEVA-Net is to investigate how to systematically generate descent domain-specific prompts for better view aggregation enhancement, such as leveraging large language models.

% \begin{acks}
% \end{acks}

\bibliographystyle{ACM-Reference-Format}
\bibliography{OurReference}

\end{document}